\documentclass[preprint,12pt]{elsarticle}

\usepackage{amssymb}
\usepackage{amsmath}
\usepackage{comment}
\usepackage{array}
\usepackage{booktabs}
\usepackage{multirow}
\usepackage[T1]{fontenc}
\usepackage{textcomp}
\usepackage{lmodern}
\usepackage{graphicx}
\usepackage{subcaption}

\journal{Applied Soft Computing}

\begin{document}

\title{Automated Flow Pattern Classification in Multi-phase Systems Using AI and Capacitance Sensing Techniques}

\begin{comment}
\author{Nian Ran}
\affiliation{%
  \organization{The University of Manchester}
  \city{Manchester}
  \country{United Kingdom}
  \email{r992988188@gmail.com}
}

\author{Fayez M. Al-Alweet}
\affiliation{%
  \institution{King Abdulaziz City for Science and Technology}
  \city{Riyadh}
  \country{Saudi Arabia}
  \email{falalweet@kacst.edu.sa}
  }

\author{Richard Allmendinger}
\affiliation{%
 \institution{The University of Manchester}
  \city{Manchester}
  \country{United Kingdom}
  \email{richard.allmendinger@manchester.ac.uk}
}

\author{Ahmad Almakhlafi}
\affiliation{%
 \institution{DigitalU Technologies}
 \city{Riyadh}
 \country{Saudi Arabia}
 \email{a.almakhlafi@digitalu.sa}
 }
\end{comment}
% \corref{cor1}   corresponding author
\author[manchester]{Nian Ran}
\ead{r992988188@gmail.com}

\author[kacst]{Fayez M. Al-Alweet}
\ead{falalweet@kacst.edu.sa}

\author[manchester]{Richard Allmendinger\corref{cor1}}
\ead{richard.allmendinger@manchester.ac.uk}

\author[digitalu]{Ahmad Almakhlafi}
\ead{a.almakhlafi@smartu.sa}

%% Corresponding author footnote
\cortext[cor1]{Corresponding author.}

%% Affiliations
\affiliation[manchester]{organization={The University of Manchester}, 
            city={Manchester},
            country={United Kingdom}}

\affiliation[kacst]{organization={King Abdulaziz City for Science and Technology}, 
            city={Riyadh},
            country={Saudi Arabia}}

\affiliation[digitalu]{organization={SmartU Technologies}, 
            city={Riyadh},
            country={Saudi Arabia}}


\begin{abstract}
In multiphase flow systems, classifying flow patterns is crucial to optimize fluid dynamics and enhance system efficiency. Current industrial methods and scientific laboratories mainly depend on techniques such as flow visualization using regular cameras or the naked eye, as well as high-speed imaging at elevated flow rates. These methods are limited by their reliance on subjective interpretations and are particularly applicable in transparent pipes. Consequently, conventional techniques usually achieve context-dependent accuracy rates and often lack generalizability. This study introduces a novel platform that integrates a capacitance sensor and AI-driven classification methods, benchmarked against traditional techniques. Experimental results demonstrate that the proposed approach, utilizing a 1D SENet deep learning model, achieves over 85\% accuracy on experiment-based datasets and 71\% accuracy on pattern-based datasets. These results highlight significant improvements in robustness and reliability compared to existing methodologies. This work offers a transformative pathway for real-time flow monitoring and predictive modeling, addressing key challenges in industrial applications.
\end{abstract}

\begin{comment}
\begin{CCSXML}
<ccs2012>
 <concept>
  <concept_id>00000000.0000000.0000000</concept_id>
  <concept_desc>Do Not Use This Code, Generate the Correct Terms for Your Paper</concept_desc>
  <concept_significance>500</concept_significance>
 </concept>
 <concept>
  <concept_id>00000000.00000000.00000000</concept_id>
  <concept_desc>Do Not Use This Code, Generate the Correct Terms for Your Paper</concept_desc>
  <concept_significance>300</concept_significance>
 </concept>
 <concept>
  <concept_id>00000000.00000000.00000000</concept_id>
  <concept_desc>Do Not Use This Code, Generate the Correct Terms for Your Paper</concept_desc>
  <concept_significance>100</concept_significance>
 </concept>
 <concept>
  <concept_id>00000000.00000000.00000000</concept_id>
  <concept_desc>Do Not Use This Code, Generate the Correct Terms for Your Paper</concept_desc>
  <concept_significance>100</concept_significance>
 </concept>
</ccs2012>
\end{CCSXML}
\end{comment}

%\ccsdesc[500]{Do Not Use This Code~Generate the Correct Terms for Your Paper}
%\ccsdesc[300]{Do Not Use This Code~Generate the Correct Terms for Your Paper}
%\ccsdesc{Do Not Use This Code~Generate the Correct Terms for Your Paper}
%\ccsdesc[100]{Do Not Use This Code~Generate the Correct Terms for Your Paper}

\begin{keyword}Multiphase flow \sep Flow pattern classification \sep Artificial intelligence \sep Capacitance sensors \sep Two-phase flow \sep Machine learning \sep Flow visualization \sep Neural networks
\end{keyword}

\begin{figure*}[!t] % This environment makes the Nomenclature section span both columns
    \section*{Nomenclature}
    \subsection*{Latin Symbols}
    \begin{itemize}
        \item \( C(t) \) - Capacitance of the sensor as a function of time [F or V]
        \item \( \mathcal{F}\{f\} \) - Fourier transform of the function
        \item \( f \) - Frequency [Hz]
        \item \( f_s \) - Sampling frequency [Hz]
        \item \( f_c \) - Maximum effective frequency of the fluctuating signal [Hz]
        \item \( K \) - A constant typically chosen between 2 and 3 to ensure adequate sampling, prevent aliasing, and improve signal reconstruction accuracy
        \item \( u_{\text{OS}} \) - Superficial oil velocity [m/s]
        \item \( u_{\text{GS}} \) - Superficial gas velocity [m/s]
    \end{itemize}

    \subsection*{Greek Symbols}
    \begin{itemize}
        \item \( \theta \) - Pipe inclination [\(^\circ\)]
        \item \( \rho_G \) - Gas density [kg/m\(^3\)]
        \item \( \rho_L \) - Liquid density [kg/m\(^3\)]
    \end{itemize}

    \subsection*{Abbreviations}
    \begin{itemize}
        \item ANN - Artificial Neural Network
        \item DFT - Discrete Fourier Transform
        \item ECT - Electrical Capacitance Tomography
        \item FFT - Fast Fourier Transform
        \item PDF - Probability Density Function
        \item PIV - Particle Image Velocimetry
        \item PSD - Power Spectral Density
        \item CMD - Capacitance Measurement Device
        \item SM-02 - Capacitive transducer
    \end{itemize}
\end{figure*}


\maketitle

\section{Introduction}\label{intro}
\begin{figure}[h]
\centering
\includegraphics[width=0.4\textwidth]{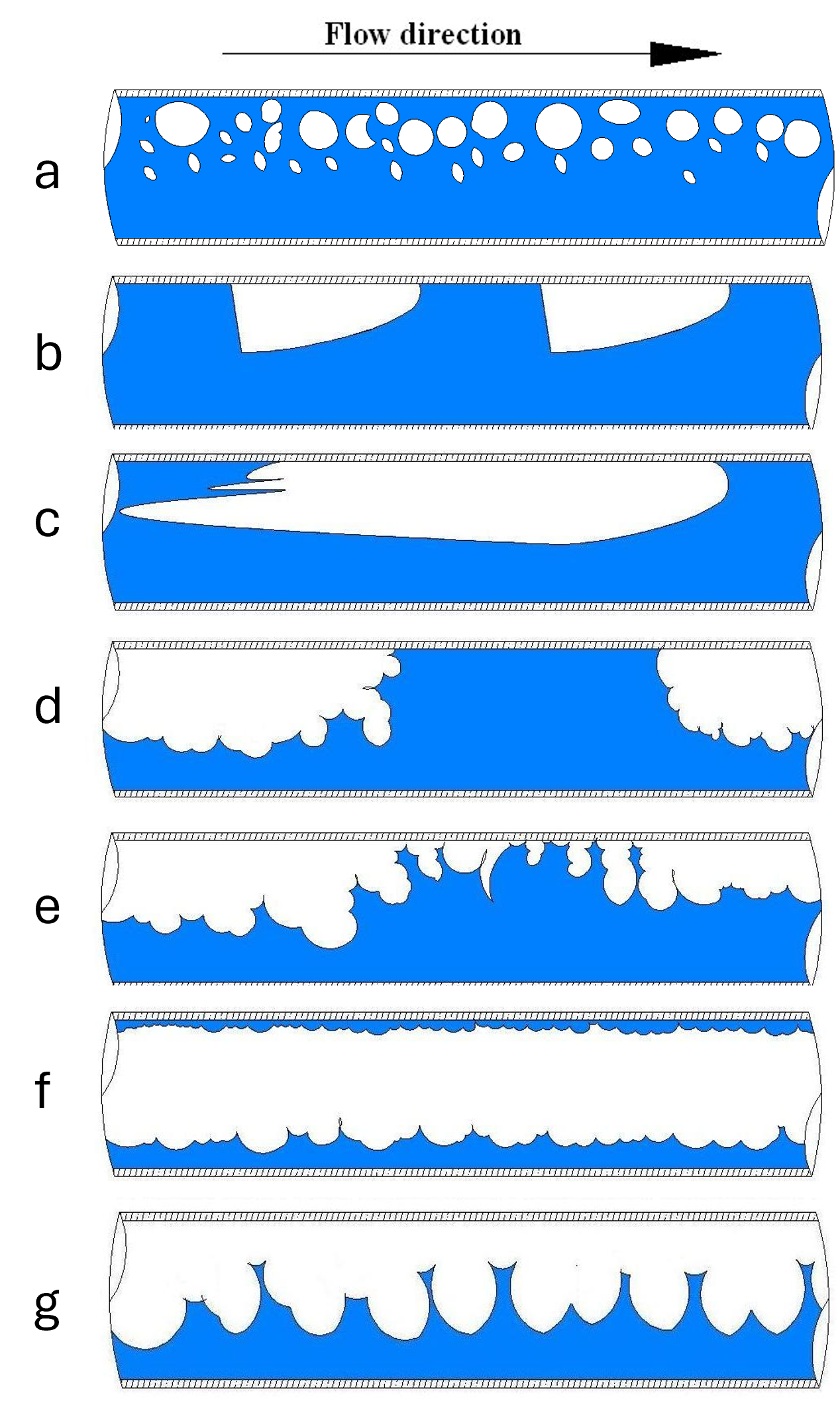}
\caption{Illustration of common two-phase flow patterns observed in pipelines. From top to bottom: (a) Dispersed Bubble Flow, characterized by small gas bubbles distributed within the liquid phase; (b) Plug Flow, featuring larger gas bubbles separated by liquid plugs; (c) Elongated Bubble Flow, where bullet-shaped gas bubbles dominate; (d) Slug Flow, alternating liquid slugs and gas pockets; (e) Churn Flow, a chaotic mixture of frothy liquid and gas; (f) Annular Flow, with a liquid film along the pipe walls and gas flowing centrally; and (g) Stratified Flow, where gas and liquid phases are clearly separated.}
\label{fig:Patterns_vis}
\end{figure}

The term \textit{multiphase flow} refers to the concurrent movement of multiple phases, including gases, liquids, or solids. \textit{multi-component} flow describes situations where these phases consist of distinct chemical substances, allowing for the categorization of different types of multiphase flow. The first type involves a single component in two phases, such as steam and water. The second type encompasses the simultaneous flow of two components, like oil and gas, each in a separate phase. The third type combines two immiscible liquids, like oil and water, known as single-phase, two-component flow. Generally, gases tend to mix unless there is a significant density difference and limited inter-phase interactions~\cite{wallis1969one, soo1990multiphase, kleinstreuer2003two, hewitt1969phase}. Figure \ref{fig:Patterns_vis} illustrates the diverse flow patterns commonly encountered in two-phase flow systems, highlighting the complexity and variability of flow regimes that require accurate classification for effective system design and operation ~\cite{al2008development}.

Multiphase flow is prevalent in various industrial applications, including chemical and paper-making plants, nuclear reactors, steam generators, and cooling systems. In the petroleum industry, it occurs in wellbores where natural gas, crude oil, water, and solids coexist and flow together. Offshore drilling involves transporting these components through pipelines to onshore refiners. In the chemical industry, two-phase flow is common when gas and liquid are introduced into reactors, facilitating heat and mass transfer or chemical reactions. Examples can be seen in evaporators, condensers, and distillation units~\cite{wallis1969one,soo1990multiphase,kleinstreuer2003two}.

The behavior of multiphase flow is more complex than that of a single-phase flow because of distinct fluid property interfaces. Two-phase flow exhibits various configurations within pipes, known as flow regimes or patterns, which differ from laminar and turbulent flow in the single-phase flow. While the classification system for single-phase flow is well-established, no consensus exists regarding two-phase flow patterns. At high flow rates, two-phase flow becomes chaotic and difficult to accurately define~\cite{kleinstreuer2003two,hewitt1969phase,songsiri2004tow}.

Common patterns in two-phase flow include dispersed bubble, plug, elongated bubble, slug, churn, annular, and stratified configurations. Each pattern has unique characteristics and influences the overall flow behavior. However, categorization is not standardized and subjective interpretations complicate the establishment of a universal classification system. Understanding flow patterns is vital, as they affect parameters such as liquid hold-up and pressure drop, necessitating analysis for prediction and control~\cite{albion2007flow,barnea1986transition,barnea1987unified,barnea1990effect,barnea1980flow,barnea1980flow,barnea1982bflow,barnea1980bgas,barnea1985gas,barnea1986bflow,barnea1993model}.

The study of two-phase flow is complex due to multiple variables, including slip between phases and varying interfaces, leading to higher pressure losses than in single-phase flow. Factors like pipe diameter, orientation, and flow rates significantly impact flow patterns. Therefore, models must consider these variables and designers should predict flow patterns based on available information. Although there are various measurements to determine the holding capacity and pressure drop of the liquid, they may not be reliable for inclined flow~ \cite{barnea1986bflow,taitel1976model,taitel1980modelling,abduvayt2003effects}.

The subjective nature of interpreting flow patterns complicates their description, particularly at high flow rates, where visualization is impractical. Efforts are underway to develop objective flow pattern detection techniques for opaque pipes. Identifying flow patterns in advance is crucial for measuring two-phase flow and helps engineers analyze problems effectively. Various ideas and advances have emerged to address the determination and classification of flow patterns~\cite{mi1998vertical,van1993spatial,hernandez2006fast,xie2003flow,xie2004artificial}.

The classification and definition of flow patterns are crucial in two-phase flow, as they directly impact key parameters, such as liquid hold-up and pressure drop. However, a notable lack of artificial intelligence (AI) methodologies exists for flow pattern classification using capacitance sensor data. To address this, a flow platform capable of replicating all categories of flow patterns will be constructed. This platform will utilize a simple capacitance sensor for data collection, which will be analyzed through AI algorithms, combined with high-speed cameras and visual feedback. The need for research on AI applications in flow pattern classification is essential to bridge this gap. Accurate classification of flow patterns is critical for improving the design and operation of multiphase flow systems, improving efficiency and safety. By reducing subjectivity in flow pattern identification, more consistent and reliable predictions of system behavior can be achieved, which is essential for applications ranging from oil and gas pipelines to chemical reactors. Although AI offers promising solutions for automating and improving the accuracy of flow pattern classification, existing studies have mainly focused on data from different sensors. However, a gap remains in the use of AI with simple capacitance sensors to achieve comprehensive flow pattern classification.

The key contributions of this study are summarized as follows, highlighting the advancements made in addressing the challenges of real-time and accurate flow pattern classification in multiphase systems. 

\begin{enumerate}
    \item \textbf{Development of an AI-Based Classification System}: A novel AI framework specifically designed to analyze data from a simple capacitance sensor is introduced to classify flow patterns.
    \item \textbf{Integration of High-Speed Cameras and Visual Feedback}: By combining visual data with sensor readings, the accuracy and robustness of the classification process are enhanced.
    \item \textbf{Comprehensive Testing Platform}: A flow platform capable of replicating various flow patterns across multiple phases is constructed, allowing for extensive validation of the proposed system.
    \item \textbf{Advancement in Predictive Modeling}: The study pioneers the application of AI in the classification of flow patterns with capacitance sensors, demonstrating the potential for scalable predictive modeling in multiphase flow systems.
\end{enumerate}

The remainder of this paper is structured as follows. Section~\ref{LitReview} presents a review of the literature on flow pattern classification and the state-of-the-art in the application of AI in this field. Section~\ref{method} describes the AI models (and their parameters) that we investigate in this study for flow pattern classification and the data acquisition and pre-processing employed. Section~\ref{results} presents and analyzes the experimental results, and Section~\ref{conclusion} concludes the article and discusses future research. 

\section{Literature Review}\label{LitReview}
Extensive studies on two-phase flow patterns have been conducted since the early 1960s. Experiments revealed that flow patterns are influenced by flow rate, showing that two-phase flow is fundamentally distinct and more intricate than single-phase flow. These distinct configurations, known as flow regimes or flow patterns, depend on variables such as gas and liquid flow rates, fluid properties (density, viscosity, and surface tension), and geometrical factors (conduit shape, diameter, and inclination). A significant distinction is the incomplete or partial conversion of potential energy to pressure energy in downward flow, whereas in upward flow, much of the potential energy can be converted to kinetic energy, depending on the flow pattern \cite{hewitt1969phase,taitel1976model,mandhane1974flow,matsui1984identification,matsui1986identification,kokal1989aexperimental,ekberg1999gas,spedding1993flow,hong1997effect}.

Differentiating between flow patterns is crucial to understand multiphase flow structures. Visual inspection in transparent pipes is the common method, allowing automatic visual representation and analysis. However, standardizing flow pattern classification is challenging due to difficulties in accurate identification. Researchers face obstacles in categorizing flow patterns, which leads to discrepancies in designation by different methods. These issues stem from diverse detection methods and instruments. In addition, flow patterns may not develop fully at the instrument location, leading to observational discrepancies. Evaluating and comparing data from different experiments is challenging due to various instruments, methods, and undefined calibration procedures. Recent categorizations have been developed and accepted, yet naming discrepancies persist due to language variations, subjective naming practices, and lack of consensus, complicating consistency and data analysis in studies~\cite{songsiri2004tow,abduvayt2003effects,hernandez2006fast,kokal1989aexperimental,ekberg1999gas,spedding1993flow,bousman1996gas,elperin2002flow,somchai2006flow}.

Despite various contributions to understanding flow patterns, a consensus on a uniform classification and nomenclature remains elusive. Consequently, the focus will shift to widely accepted flow patterns, presenting alternative names when relevant. Each pattern will be briefly defined to highlight its distinctive form and mechanism. Dispersed bubble flow is characterized by small, discrete gas bubbles suspended in a predominantly liquid flow. In this scenario, the bubbles tend to rise to the top at low gas flow rates but become uniformly dispersed when the gas flow rates increase. This flow pattern is also referred to as ``bubbly flow''. Next is plug flow, which involves the clustering of small gas bubbles that merge to form larger, bullet-shaped bubbles. This phenomenon is sometimes known as ``cap-bubbly flow''. Elongated bubble flow features gas bubbles that take on a streamlined shape, resulting in an intermittent flow pattern. In this case, the rear end of a bubble may detach and be captured by the following bubble, creating a dynamic flow environment. Slug flow consists of sections where the gas phase nearly fills the entire cross-section of the pipe, alternating with liquid sections. This pattern is most common in horizontal and upward flows~\cite{soo1990multiphase,kleinstreuer2003two,hewitt1969phase}. Further complicating the behavior is slug froth flow, which is marked by highly frothy liquid slugs created through turbulence and mixing. This pattern, also referred to as ``churn flow'' or ``slug-churn'', often results in significant pressure drops~\cite{soo1990multiphase,kleinstreuer2003two}. In annular flow, liquid forms a film that coats the pipe wall, while gas flows through the center. This configuration can lead to the development of unstable waves as a result of the mixing of the two phases. Finally, stratified flow is defined by a clear separation between the liquid at the bottom of the pipe and the gas at the top. This flow pattern can be divided into two categories based on the gas flow rates: stratified smooth and stratified wavy flow~\cite{al2020bsimplified,al2008development,cheremisinoff1979stratified}.

In multiphase flow, various devices, instruments, and techniques are used to define and distinguish between different flow patterns. Some commonly used ones include:

\begin{enumerate} 
    \item \textbf{Flow Visualization Techniques}: Transparent pipes or flow channels with dyes or tracers to observe flow regimes~\cite{al2020systematic,al2020bsimplified,tsubone2001effects,almutairi2020ect}.
    
    \item \textbf{High-Speed Imaging}: Captures the transient behavior of flows, allowing identification and classification of patterns~\cite{somchai2006flow,ito2001flow,bennett2006frequency,xia1996two,clarke2001study,van2001evolution}.
    
    \item \textbf{Pressure Drop Measurements}: Pressure sensors determine gradients and identify patterns based on pressure drop characteristics~\cite{matsui1984identification,matsui1986identification,spedding1993flow,samways1997pressure,li2002experiment}.
    
    \item \textbf{Electrical Conductivity Measurements}: Weak electrical currents identify changes in liquid hold-up and differentiate flow patterns~\cite{hernandez2006fast,lamb1960measurement,van1985void,liu1993bstructure,jin2008design}.
    
    \item \textbf{Acoustic Techniques}: Ultrasonic sensors or acoustic impedance probes detect distinctive acoustic signatures of flow patterns~\cite{albion2007flow,xu2000acoustic,chung2004sound,gadiyaram2005acoustic}.
    
    \item \textbf{Optical Probes}: Quantify flow characteristics, such as velocity and turbulence, to identify patterns~\cite{yoon2006gas,polonsky1999relation,tu2002methodology,nydal1992statistical,andreussi1993void,martin2000slug}.
    
    \item \textbf{Gamma-Ray Transmission}: Measures attenuation profiles to differentiate flow patterns~\cite{xie2003flow,tortora2004capacitance,simons1995imaging,bieberle2006evaluation,hanus2022investigation}.
    
    \item \textbf{Electrical Capacitance Sensor}: Two-electrode capacitor detects flow patterns by analyzing electrical signal changes caused by material flow~\cite{williams1995process,abouelwafa1980use,geraets1988capacitance,tollefesn1998capacitance}.
    
    \item \textbf{Tomographic Methods}: Reconstruct cross-sectional images of flow behavior using data from multiple sensors around the system, including electrical, optical, gamma-ray, and X-ray tomography techniques~\cite{simons1995imaging,gamio2005visualisation,makkawi2002fluidization,reinecke1997multielectrode,yang2004adaptive,warsito2001measurement,dong2003application,ma2001application,lee2014electrical}.

\end{enumerate}

Data analysis methods play a vital role in understanding two-phase flow patterns, which are essential for various industrial processes and engineering applications. Using analytical techniques, valuable insights can be gained into two-phase flows, which facilitate informed decision making and process optimization. Among the methods used are flow pattern maps, image analysis, statistical analysis, neural networks, pattern recognition, and artificial intelligence (AI), each of which contributes to understanding the characteristics of the flow in two phases.

Flow pattern maps serve as graphical representations that classify different flow patterns based on experimental observations. Using dimensionless parameters, such as the Weber number and superficial velocities, these maps define boundaries between flow behaviors, allowing comparison between conditions~\cite{songsiri2004tow,barnea1986transition,mandhane1974flow,spedding1980regime,stanislav1986intermittent}. Image analysis techniques involve the capture of visual data with high-speed cameras, processing these data to extract information on bubble size, velocity, void fraction, and shape, helping to identify flow patterns~\cite{al2020systematic,al2020bsimplified,al2008development,almutairi2020ect}.

Statistical analysis employs methods such as cluster analysis, principal component analysis (PCA), and discriminant analysis to explore experimental data. Flow patterns can be classified according to statistical properties, offering quantitative classification means~\cite{al2020systematic,van2001evolution,nydal1992statistical,matsumoto1984statistical,chakraborty2020characterisation,ameel2012classification,yang2017application,nie2022image,trafalis2005two}. 

There is a body of research that has explored the application of AI techniques to classify flow patterns, such as support vector machines (SVM), decision trees, and artificial neural networks (ANNs)~\cite{nie2022image,trafalis2005two,sun2023comparative,li2023gas,howard2007pattern,mi1998vertical,hernandez2006fast,xie2003flow,xie2004artificial,chakraborty2020characterisation,yang2017application}. These methods consider input variables such as pressure drop, void fraction, and flow rate to capture complex relationships in flows. Moreover, AI, in particular deep learning, has also been used in combination with computer vision to detect flow patterns based on images of flows. In signal processing, AI algorithms have been tested on sensor data to classify flow patterns; data fusion has also been used to integrate information from multiple sources, enhancing the understanding of two-phase flow behaviors~\cite{klein2004sensor,hall1997introduction,zhang2014data,barbariol2020sensor}. However, previous work requires either high-speed camera or many sensors to obtain input data which are expensive and difficult to deploy in practice. On top of that, the exploration of machine learning models and the number of flow pattern types in those work remains preliminary. In this work, we only utilize data from two sensors and 5 second time-series signals to classify 7 types of flow patterns by 1D convolutional neural networks, compared with other machine learning models.

It should be noted that the methods reviewed in this section are often used in combination to provide a more comprehensive analysis of the two-phase flow behavior. The choice of method depends on the available data, the specific research objectives, and the desired level of accuracy and detail in flow pattern classification.

\section{Methodology}\label{method}
This section presents a comprehensive account of the construction of a flow rig designed to generate a complete spectrum of two-phase flow patterns in three specific pipe orientations: 0°, 10°, and 20°. The objective of this experimental phase is twofold: to observe and document the entire range of flow patterns in great detail, and to comprehend the transitional behavior between different flow patterns. Identification and analysis of flow patterns were carried out using a high-speed camera. Multiple photographs were taken for each flow pattern, which depicted its physical and mechanical characteristics. Concurrently, experimental data on flow pattern transitions within the oil and gas flow system were acquired through real-time monitoring by the high-speed camera and capacitance sensor. By combining visual observations and recorded data, a comprehensive understanding of the flow pattern transitions in the oil and gas flow system was obtained.

\subsection{Experimental Setup}
\begin{figure*}[h]
\centering
\includegraphics[width=0.8\textwidth]{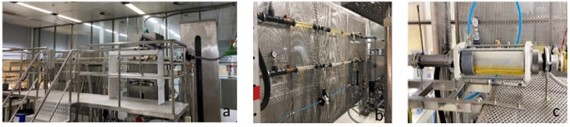}
\caption{(a) Photographic view of flow rig, (b) Gas injection panel and flow speed control devices, (c) Gas-liquid mixer.}
\label{fig:setup}
\end{figure*}

\begin{figure*}[h]
\centering
\includegraphics[width=0.8\textwidth]{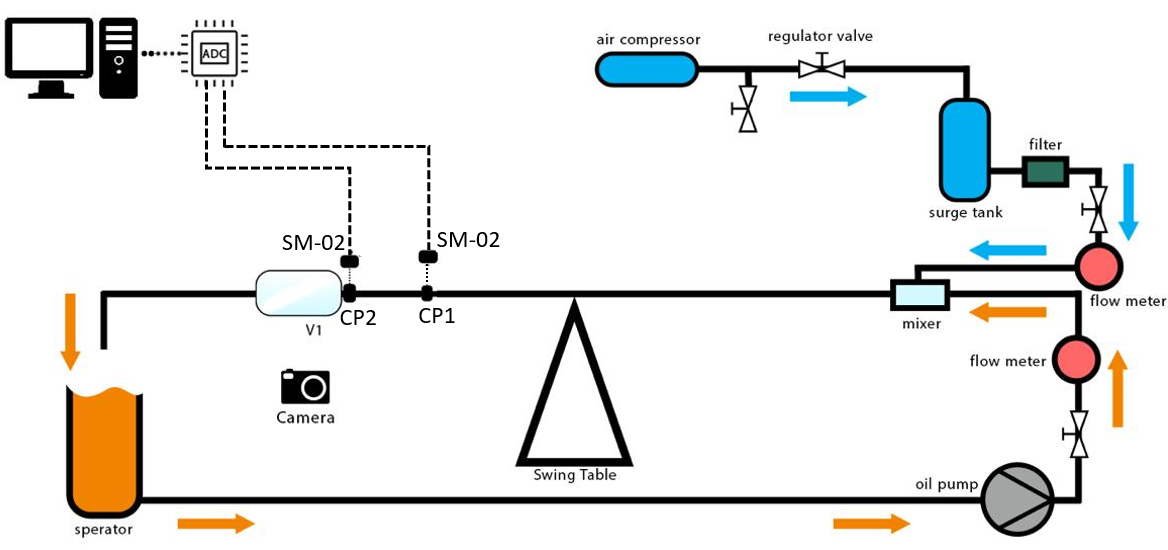}
\caption{ A Schematic layout of the flow rig.}
\label{fig:setup2}
\end{figure*}

The flow loop was designed to function as an open recirculating system, with the aim of efficiently mixing different phases and promoting the formation of diverse flow patterns. Constructed using a transparent acrylic PVC pipe, the loop spanned a total length of 25 meters, with an inner diameter of 36 mm and an outer diameter of 40 mm. The loop consisted of various components, including a test section, a liquid tank, a pump, an air compressor, and a swing table as shown in Figure~\ref{fig:setup} (a) and Figure~\ref{fig:setup2}. (The setup is located at King Abdulaziz City for Science and Technology in Saudi Arabia). 

The test section, which was approximately 6 meters long, was made of transparent PVC R-4000 pipe with a 36 mm inner diameter, allowing easy visualization of flow behavior. Two flexible elbow pipes were incorporated into the loop's design to allow for rotation and adjustment. This rotation capability allowed the loop to be inclined at various angles, ranging from horizontal to 30° upward flow. To minimize the impact of vibrations on the system, a swing table was used to effectively support the test section.

The two-phase flow from the return section of the loop was directed into a liquid tank, which served the purpose of separating the liquid and air components after each series of experiments. This separation was necessary to prevent air from entering the pump. The liquid tank had a volume of approximately 2.0 m³. To circulate the liquid through the flow loop, a gear pump with a presumed flow rate of 200 L/min was utilized. To ensure that the liquid did not flow back, a check valve was installed in the system.

An air compressor injected gas downstream of the liquid flow meter into the loop. This was accomplished by using a brass pipe, approximately 2.50 meters long with an inner diameter of 9.0 mm, connected to the compressed air line. To ensure unidirectional flow, a check valve was installed in the setup as shown in Figure~\ref{fig:setup} (b).

\subsection{Experimental Apparatus}
An experimental apparatus was meticulously designed and calibrated to ensure the successful implementation of our research. This apparatus was specifically customized to meet the precise demands of our research objectives, allowing accurate measurement and control of key variables, such as the capacitance sensor and high-resolution camera recordings.

The test section, measuring 6.0 meters in length, was fabricated using an extruded transparent acrylic tube with an internal diameter of 36.0 mm and an outer diameter of 40.0 mm. This design enables visual assessment of flow dynamics and incorporates a swinging table mechanism that modifies its inclination from 0$^\circ$ to 35$^\circ$. Additionally, flexible tubes at the edges facilitate investigations at user-defined angles. To mitigate vibration impacts, the test section was supported by two jacks. Fluid supply involved extracting oil from a 2 m$^3$ tank, which was pumped through a PVC pipe using a gear pump (Model GL-50-10). A check valve was installed to prevent backflow.

\begin{figure}[h]
    \centering
    \begin{minipage}[t]{0.45\textwidth}
        \centering
        \includegraphics[width=\textwidth, keepaspectratio]{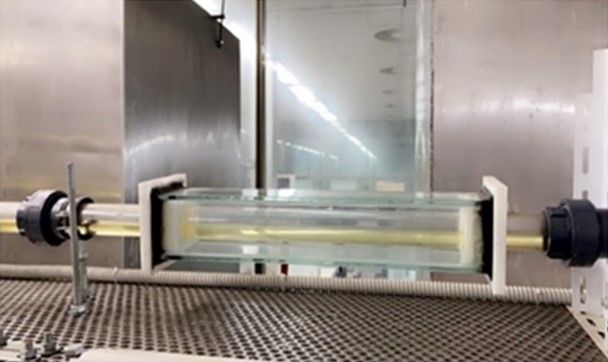}
        \caption{Photographic viewing box.}
        \label{fig:viewbox}
    \end{minipage}
    \hfill
    \begin{minipage}[t]{0.45\textwidth}
        \centering
        \includegraphics[width=\textwidth, keepaspectratio]{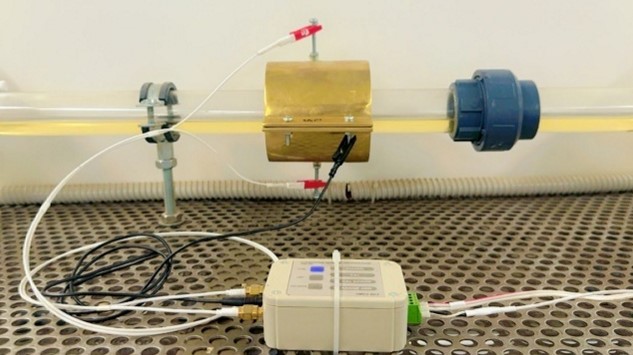}
        \caption{Capacitance sensor clamped on the test-section pipe and its electronic measurement device.}
        \label{fig:sensor}
    \end{minipage}
    %\caption{Combined Figure: Viewing box and Capacitance sensor setup.}
\end{figure}

Gas injection was facilitated by positioning an air compressor downstream of the liquid flow meter. Effective homogenization of oil and gas was achieved through an innovative mixer device measuring 28.0 cm. This mixer featured an external pipe of Perspex and an internal brass pipe, both designed with numerous small holes. The inner pipe, matching the test section diameter, was separated from the outer pipe by a 1.8 cm gap and contained approximately 100 staggered holes, with each hole measuring 1 mm in diameter and spaced appropriately for smooth gas entry, as shown Figure~\ref{fig:setup} (c). Oil flowed in the axial direction while air was introduced into the space between the pipes, entering the test section through the holes. This cyclical process enabled the oil and gas to return to the oil tank, with the gas naturally separating and evaporating into the environment.

The experimental apparatus was illuminated for visual observation and recording of flow patterns. A transparent rectangular box filled with water was incorporated to mitigate image distortions caused by pipe curvature and cool down the heating effect from illumination, as shown in Figure~\ref{fig:viewbox}. Flow characteristics were captured using an Analogue High-Speed Video system (CR4000 $\times$ 2, Optronis GmbH, Kehl, DE), which acquired 500 frames per second for durations of 10-20 seconds, achieving a resolution of 4 megapixels. The camera was equipped with an F mount and a Nikon 50 mm f/1.8D lens, ensuring uniform lighting during recordings via an LED assembly.

\begin{figure*}[h]
\centering
\includegraphics[width=0.8\textwidth]{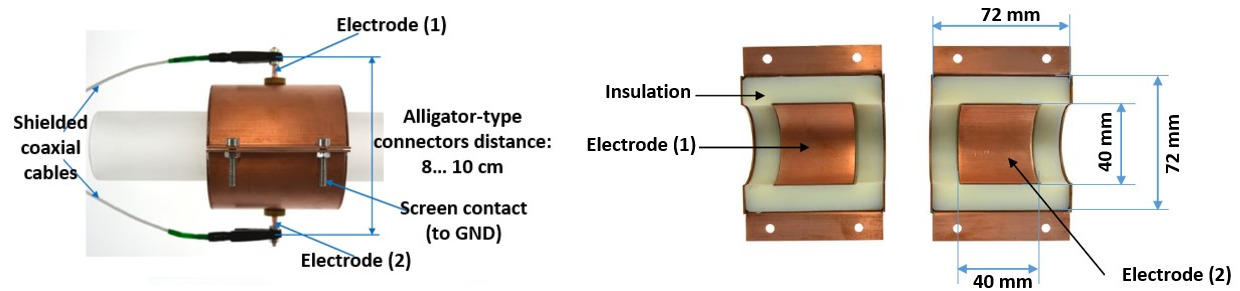}
\caption{The capacitance sensor utilized in the experiments, along with a diagram illustrating its geometry.}
\label{fig:sensor_geom}
\end{figure*}

The capacitance sensor was installed 5 meters from the inlet and securely clamped around the test section. It was designed to provide accurate flow pattern detection using two opposing electrodes filled with acrylic insulation material to minimize electrical interference. The positioning of the excitation electrode at the top of the pipe ensured precise detection of the flow pattern~\cite{al2020bsimplified, al2020systematic}. This configuration effectively mitigated issues such as erosion and disruption of flow patterns. The capacitance sensor was connected to a dedicated electronic capacitance measurement device, with outputs transmitted to an analogue-to-digital converter (ADC) via a shielded cable designed to minimize noise interference, as shown in Figure~\ref{fig:sensor}. Both the sensor and electronic devices were calibrated to operate within a capacitance range of 0 to 100 pF.

Capacitance measurements were performed using a device known as the Capacitance Measurement Device (CMD), equipped with a burst mode charge-transfer converter, as shown in Figure~\ref{fig:sensor_geom}. The CMD features a Pulse Width Modulation (PWM) output with specialized filters for gas/liquid flow applications. It has eight-bit resolution and two calibration inputs for span calibration. The CMD automatically compensates for cable capacitance, enhancing measurement accuracy. Its control panel allows for easy calibration with LED indicators available for displaying status and errors~\cite{li2002accurate, toth1997design, ferry1997design}. Capacitive sensors detect changes in electrical capacitance due to physical property variations, typically measuring in picofarads. However, measuring such low capacitance accurately can be challenging due to external interferences, including noise and parasitic capacitances. The CMD addresses these issues and delivers accurate measurements, supported by previously established methodologies.

The study employed a Capacitive Sensor Transducer capable of measuring capacitance values up to 24 pF, with calibration within any subrange from 0 to 24 pF. Utilizing circuitry devices and an analog low-pass filter in conjunction with PWM output, small capacitance values were converted into direct current voltage signals ranging from 0 to 5 V. This configuration reduced power consumption and noise interference while improving circuit sensitivity and stability~\cite{ferry1997design, kollataj2008multi, smith1997scientist}. For a more comprehensive description of the methodology, refer to the Appendix~\ref{app1}.

\subsection{Methodology and Experimental Approach}
A sequence of experimental trials were conducted to examine the various flow patterns and their evolutions that could arise in a two-phase flow occurring within a pipeline. The experimental setup involved acquiring room temperature air, characterized by a density of $\rho(G) = 1.204$ kg/m$^3$, from the air compressor line located in the laboratory of King Abdulaziz City for Science and Technology. Subsequently, the air was subjected to a pressure reducer to ensure a consistent flow rate. The flow rate of the air was meticulously determined using a gas ball flow meter boasting an accuracy of $\pm0.6\%$. In addition, a mass flowmeter (FMA 1700/1800 model, Omega, Norwalk, CT, USA) with a precision of $\pm0.1\%$ was also used for precise measurement. Additionally, mineral oil with a density of $\rho(L) = 850$ kg/m$^3$ was conveyed from an oil tank at room temperature and atmospheric pressure using a gear pump. The oil flow rate was regulated by modifying the speed of the pump motor and was precisely gauged by a turbine flowmeter (FTB790-Omega) boasting an accuracy of $\pm0.2\%$. 

In order to examine the impact of varying flow rates, the air flow rate was systematically modified while maintaining a constant oil flow rate at a predetermined value. The superficial velocities of both the gas and liquid were regarded as controlled variables. The range of superficial gas velocities tested ranged from 0.501 to 5.0 m/s, with increments of 0.25 m/s. Similarly, the range of superficial liquid velocities tested ranged from 0.106 to 1.06 m/s, with increments of 0.106 m/s. To monitor flow conditions, several pressure and temperature gauges were installed within the flow rig. The liquid temperature was carefully regulated to remain near a constant value of 20$^{\circ}$C (equivalent to room temperature) throughout each trial. This temperature was measured using a thermometer placed within the tank.

After all the experimental parameters were configured, such as the angle of the test section, calibration of the capacitance sensors, superficial liquid and gas velocities, and the temperature of the liquid, the flow loop was left to attain a state of equilibrium. This process usually took approximately eight minutes. Data collection encompassed simultaneous video recording, visual observations, note-taking, and recording the data from the capacitance sensors regarding flow patterns.

\subsection{Data Acquisition and Pre-processing}
\begin{figure}[h]
    \centering
    \begin{minipage}[t]{0.48\textwidth}
        \centering
        \includegraphics[width=\textwidth, keepaspectratio]{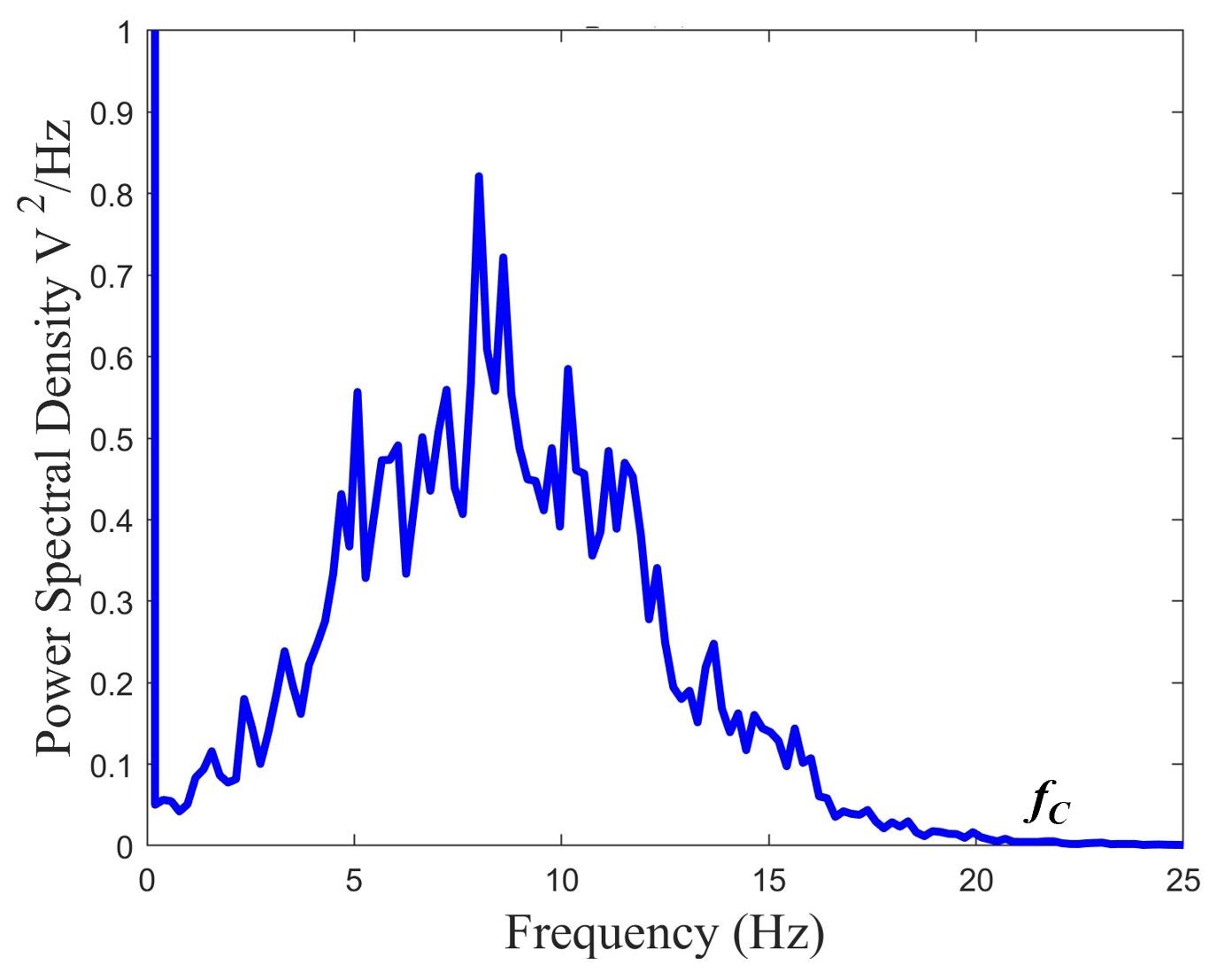}
        \caption{PSD analysis of the capacitance signal with sample size 8192, window size 31, and oil/gas superficial velocities of 0.61/0.2 m/s and inclination of 20$^{\circ}$C.}
        \label{fig:psd}
    \end{minipage}
    \hfill
    \begin{minipage}[t]{0.48\textwidth}
        \centering
        \includegraphics[width=\textwidth, keepaspectratio]{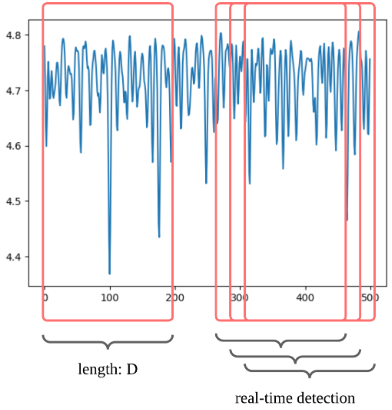}
        \caption{Data sampling example.}
        \label{fig:sampling}
    \end{minipage}
    %\caption{Combined Figure: Viewing box and Capacitance sensor setup.}
\end{figure}

Data acquisition was performed using a DI-4108-E Analog-to-Digital Converter (ADC), which converted the capacitance sensor output into digital signals. The ADC allowed for programmable ranges per channel of $\pm200$, $\pm500$ mV, $\pm1$, $\pm2$, $\pm5$, $\pm10$ V full-scale, ensuring compatibility with computer input. With a resolution of 24 bits, the ADC provided precise measurements and an accurate data representation. Its maximum sampling rate of 200 kHz facilitated the capture and recording of rapidly changing signals. The acquired digital values were processed by a computer program that continuously saved data blocks in a memory buffer. After calibration, the data was written to a CSV file. The data, along with their corresponding sampling times, were stored in an Excel file for subsequent analysis using MATLAB. The DI-4108-E ADC, with its high resolution, fast sampling rate, and compatibility with diverse applications, is well-suited for scientific research, industrial automation, control systems, and data acquisition tasks. However, a potential challenge arises regarding the impact of the sampling rate on signal fidelity during graph plotting and the generation of an accurate replica of the original signal. To address this concern, it is known that every signal possesses a frequency spectrum, which can be determined using the fast Fourier transform. According to the established definition by Schwartz~\cite{schwartz1976measurement}, the Fourier transform of a time series is represented as follows:
\begin{equation}
F(w) = \int_{-\infty}^{\infty} f(t) e^{-iwt} dt
\end{equation}
The power spectral density (PSD) of the capacitance signal was estimated using Fourier transform techniques. Our studies, conducted in collaboration with~\citet{al2020systematic, alalweet2025time}, are notable for their comprehensive exploration of calculating, deriving, and plotting the power spectral density (PSD). These works provide detailed methodologies and insights into PSD analysis, which are crucial for understanding signal characteristics in various flows patterns. The MATLAB program used 8192 data points and 31 overlapping windows to obtain the FFT spectrum. Figure~\ref{fig:psd} illustrates a representative power spectrum. The maximum frequency component of the signal was hypothesized to be $f_c$. From Figure~\ref{fig:psd} it is evident that the PSD decreases at higher frequencies and becomes negligible at $f_c$. According to sampling theory, the sampling frequency should be at least twice the highest frequency component for accurate signal reproduction~\cite{bendat1966measurement}. In simpler terms, the sampling frequency can be determined as follows:
\begin{equation}
f_s = K f_c
\end{equation}
In this study, the most efficient frequency of the oscillating signal is denoted as $f_s$ , with $K$ typically assumed to be between two and three. The precision and accuracy of estimations and computations are expected to improve as the sampling size increases, subject to limitations imposed by storage and processing capacities of the hardware. To achieve a reasonable balance, a sampling rate of 100 Hz was selected.

\subsection{Artificial Intelligence Models}
\subsubsection{Data preparation}
\label{sec:data_preparation}
High-quality and sufficient data are crucial for supervised learning, including traditional machine learning models such as decision trees and kernel methods, as well as deep learning. Thus, before applying AI, it is essential to construct a dataset for both training and evaluation. The modality and dimensions of the data also significantly impact the outcomes.

The data from capacitance sensors are temporal sequences with a sampling frequency of 100 Hz. Based on experiments on capacitance data done by Al-Alweet et al.\cite{al2020systematic,al2020bsimplified}, where 5-second intervals are typically chosen to visualize different flow patterns, we have set our sampling duration to 5 seconds as well, resulting in 500 dimensions per sample. To establish the ground truth data, we identify the experimental conditions for each flow pattern following the charts and tables provided in our previous work~\cite{al2020systematic} (see also Section~\ref{fig:Patterns_vis} and Figure~\ref{fig:Patterns_vis} for a description). Each experiment was conducted at varying superficial liquid and gas velocities, as well as different inclination angles, with specific velocity and angle ranges corresponding to each flow pattern. The data of each experiment commonly consist of 20,000 values, which represent 200 seconds of data at a 100 Hz sampling rate.

Our sampling strategy for building the dataset is divided into two approaches: experiment-based and pattern-based. In the experiment-based approach, 80\% of the data points from each experiment are allocated to the training dataset, while the remaining 20\% are reserved for evaluation. In contrast, the pattern-based approach involves randomly and uniformly selecting 80\% of the experiments for each flow pattern to generate training data, with the remaining 20\% used for evaluation. These two strategies are designed to assess both in-distribution and out-of-distribution prediction capabilities because each experiment is conducted with a combination of non-repetitive velocity and inclination angle.

To create individual data samples, we use a sliding window technique to randomly select 500 data points from the experimental data, as illustrated in Figure~\ref{fig:sampling}, where $D=500$. This approach enables the model to be trained for real-time detection. For each flow pattern, we generate 20,000 training samples and 5,000 evaluation samples, amounting to a total of 140,000 training samples and 35,000 testing samples for balanced training. The specific ranges for inclination angles, as well as superficial gas and liquid velocities for each flow pattern, are detailed in Table~\ref{tab:file-list}.

\begin{table}[h]
\centering
\footnotesize
\begin{tabular}{>{\centering\arraybackslash}m{2cm} >{\centering\arraybackslash}m{1.5cm} >{\centering\arraybackslash}m{2.5cm} >{\centering\arraybackslash}p{2.5cm}}
\toprule
Flow Pattern & Inclination & $u_{\text{GS}}$ & $u_{\text{OS}}$   \\ \midrule
\multirow{3}{1cm}{Small bubble} & $0^\circ$ & 0.000 - 0.100 & 0.675 - 1.120 \\
 & $15^\circ$ & 0.000 - 0.080 & 0.224 - 1.120 \\
 & $30^\circ$ & 0.000 - 0.100 & 0.400 - 1.120 \\ \midrule
\multirow{2}{1cm}{Plug} & $15^\circ$ & 0.127 - 0.500 & 0.530 - 1.100 \\ 
 & $30^\circ$ & 0.051 - 0.314 & 0.210 - 1.100 \\ \midrule
\multirow{3}{1cm}{Elongated bubble} & $0^\circ$ & 0.150 - 0.740 & 0.420 - 1.100 \\
 & $15^\circ$& 0.250 - 0.750 & 0.320 - 1.100 \\
& $30^\circ$ & 0.055 - 0.576 & 0.110  - 1.100 \\ \midrule
\multirow{3}{1cm}{Slug} & $0^\circ$ & 0.370 - 2.290 & 0.316 - 1.100 \\
 & $15^\circ$ & 0.700 - 2.180 & 0.120 - 1.100 \\
 & $30^\circ$ & 0.470 - 2.860 & 0.110 - 0.950 \\ \midrule
\multirow{3}{1cm}{Churn} & $0^\circ$ & 2.110 - 3.740 & 0.425 - 1.100 \\
 & $15^\circ$ & 2.900 - 4.400 & 0.110 - 1.100 \\ 
 & $30^\circ$ & 2.000 - 4.290 & 0.100 - 1.100 \\ \midrule
\multirow{3}{1cm}{Annular} & $0^\circ$ & 4.480 - 5.000 & 0.310 - 1.100 \\
 & $15^\circ$ & 4.750 - 5.000 & 0.106 - 1.100 \\
 & $30^\circ$ & 4.000 - 5.000 & 0.110 - 1.100 \\ \midrule
Stratified wavy & $0^\circ$ & 1.240 - 3.000 & 0.100 - 0.320 \\
\bottomrule
\end{tabular}
\caption{Superficial gas and liquid velocities for each flow pattern used for dataset building, where $u_{\text{GS}}$ is the superficial gas velocity and $u_{\text{OS}}$ is the superficial liquid velocity.}
\label{tab:file-list}
\end{table}

\subsubsection{Model Choices}
In this section, we introduce the AI models applied to the data after pre-processing. To comprehensively evaluate the performance of various machine learning models and given the temporal nature of our 2D signal data, we employ models from several categories: decision trees, kernel-based models, and deep learning neural networks (NN).

For decision tree-based approaches, we use Random Forest~\cite{randomForest} and the classical Decision Tree~\cite{decisionTree}, both implemented in Scikit-learn~\cite{scikit-learn}. Decision tree methods are widely applied in handling temporal and signal data. A decision tree splits data into branches based on feature values, ultimately leading to a decision at the leaf nodes. It selects the best features at each split to optimize decision-making, whereas a Random Forest builds multiple decision trees using random subsets of the data and features. The predictions are then averaged (or a majority vote is taken) to improve accuracy and reduce overfitting. Random Forest has been successfully applied in previous studies for the identification of flow patterns in various data sources~\cite{sun2023comparative}. 

Support Vector Machine (SVM) is the primary model used in kernel-based methods. SVM works by identifying the optimal hyperplane that best separates the data into distinct classes. In 2D space, this hyperplane is a line, while in higher dimensions, it becomes a plane. The key objective of SVM is to maximize the margin between data points of different classes, specifically the distance between the nearest points (support vectors) and the hyperplane. Additionally, SVM is capable of handling non-linear classification tasks through the use of kernel functions, which map the data into a higher-dimensional space where linear separation becomes feasible. It is often paired with Principal Component Analysis (PCA) to address high-dimensional features, also for flow identification tasks~\cite{shanthi2017artificial,li2022btwo,trafalis2005two}.

In the realm of deep learning, numerous models are suitable for analyzing temporal signal data, including Multilayer Perceptrons (MLP)~\cite{nn}, Convolutional Neural Networks (CNN)~\cite{cnn}, and attention-based architectures like Transformers~\cite{transformer}. Neural networks are composed of layers of interconnected nodes (neurons) that process and transform input data. Each connection between nodes has an associated weight, and neurons apply activation functions to decide whether to propagate signals. CNNs, in particular, are adept at processing grid-like data structures, such as images, by utilizing convolutional layers that detect features like edges and textures via sliding filters (kernels) across the input. Unlike traditional models, Transformers leverage self-attention mechanisms to capture relationships between words or tokens within a sequence and use feed-forward layers to process the features identified by the attention layers. In our case, signal data can be treated as tokens segmented through sliding windows. These neural network models are equipped to handle flow identification tasks, utilizing different modalities of input data~\cite{nie2022image,li2023gas,sun2023comparative,flow-transformer}.

For each model, we run experiments with 5 different random seeds and take the average results as the final performance. The default parameters for Random Forest, SVM Classifier, and Decision Tree are used as implemented in Scikit-learn. The MLP consists of three hidden layers with 200, 100, and 200 neurons, respectively. For the 1D CNN model, we use three 1D convolutional layers followed by max pooling layers, with two fully connected layers at the end. The Transformer model is built using two Transformer Encoder Layers, implemented in PyTorch~\cite{pytorch}, with a hidden dimension of 512, followed by a fully connected head. All deep learning models are trained using the AdamW optimizer for 300 epochs, and cross-entropy loss is applied for training the neural networks. The formula for cross-entropy loss is shown below:
\begin{equation}
\label{eq:ce}
  \mathcal{L} = -\frac{1}{N}\sum^N_{i=1}\sum^{C}_{c=1}y_{i,c}log(\hat{y}_{i,c})  
\end{equation}
where $N$ is the number of samples and $C$ is the number of flow pattern categories.

\begin{figure}[h]
\centering
\includegraphics[width=0.7\textwidth]{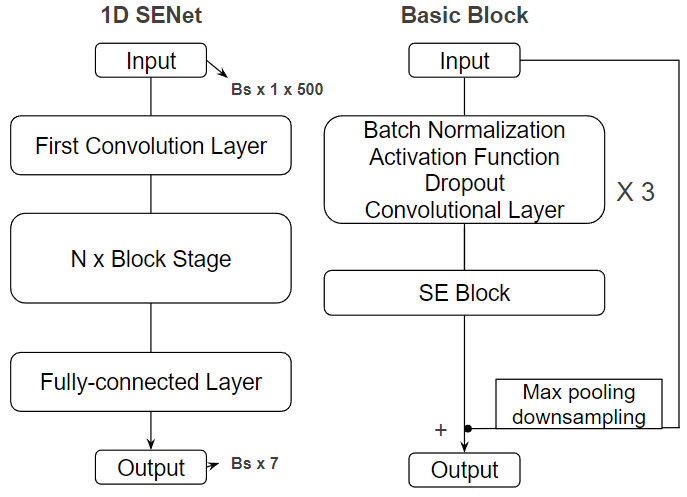}
\caption{Architecture of 1D SENet.}
\label{fig:senet}
\end{figure}

\subsubsection{SENet}
From the classification results in Section \ref{sec:exp-ai}, our most effective model is the 1D SENet that has the best performance. We will provide an overview of its architecture here. SENet is built upon ResNet, which is a milestone in machine learning for vision tasks. The core innovation of ResNet is the residual connection, which mitigates the vanishing gradient problem by adding the input to the output after a few layers, facilitating efficient and effective training of deep neural networks. This is illustrated in the right panel of Figure~\ref{fig:senet}.

SENet (Squeeze-and-Excitation Network)~\cite{senet} introduces a novel ``squeeze-and-excitation'' block, which adaptively recalibrates channel-wise feature responses. This process consists of two main steps: first, the ``squeeze'' operation compresses the spatial dimensions through global average pooling; second, the ``excitation'' step generates channel-wise weights via fully connected layers, allowing the model to enhance the most important channels and suppress less relevant ones.

Our 1D SENet consists of \textit{N} block stages, where each block stage contains multiple basic blocks. The input size is \textit{Bs x 1 x 500}, with \textit{Bs} representing the batch size and 500 referring to the data dimension, as previously mentioned. The output dimension is 7, which corresponds to the seven categories of flow patterns that the model classifies. The seven flow pattern is illustrated in figure \ref{fig:Patterns_vis}. The activation function used in SENet is Swish~\cite{swish}, a self-gated function that provides smooth, non-monotonic activation, offering improved optimization over ReLU in neural networks. The formula for Swish is as follows:
\begin{equation}
   \text{Swish}(x) = x \cdot \sigma(x), 
\end{equation}
where \(\sigma(x)\) is the sigmoid function:
\begin{equation}
  \sigma(x) = \frac{1}{1 + e^{-x}}  
\end{equation}

\section{Experimental Results and Analysis}\label{results}
Understanding the flow behavior in transparent test section tubes is essential for optimizing process performance and comprehending the underlying dynamics. The identification and classification of flow patterns in such tubes plays a crucial role in this regard. This section presents an analysis of our experimental results validating the performance of different AI methods for the classification of flow patterns. 
%In this study, a comprehensive classification of flow patterns was performed on a flow rig. This classification was based on the combination of visual observations using a high-speed camera and simultaneous monitoring using a capacitance sensor. The transparent nature of the tube allowed for real-time monitoring, which greatly facilitated the analysis and characterization of these observed flow patterns. The accuracy and reliability of flow pattern classification were improved by adopting a hybrid approach. This involved simultaneous monitoring of observed flow patterns using a capacitance sensor, which provided additional quantitative data associated with flow regimes. The precise measurements of the liquid-gas interface obtained from the capacitance sensor enabled real-time detection and tracking of flow patterns. Using the data obtained, an AI-based classification framework will be developed in subsequent sections of this study. The objective is to automate the process of classification of flow patterns employing machine learning algorithms and pattern recognition techniques. Significant improvements in pattern identification speed, efficiency, accuracy and consistency are anticipated through incorporation of this approach. The automated nature of this method will effectively mitigate the influence of human bias and subjectivity commonly encountered when relying solely on visual observations. Using machine learning algorithms and quantitative data from the capacitance sensor, the objectivity and reliability of the flow pattern classification process will be enhanced.

\subsection{Flow Patterns Observed on the Experimental Rig}
\begin{figure*}[t]
\centering
\includegraphics[width=0.8\textwidth]{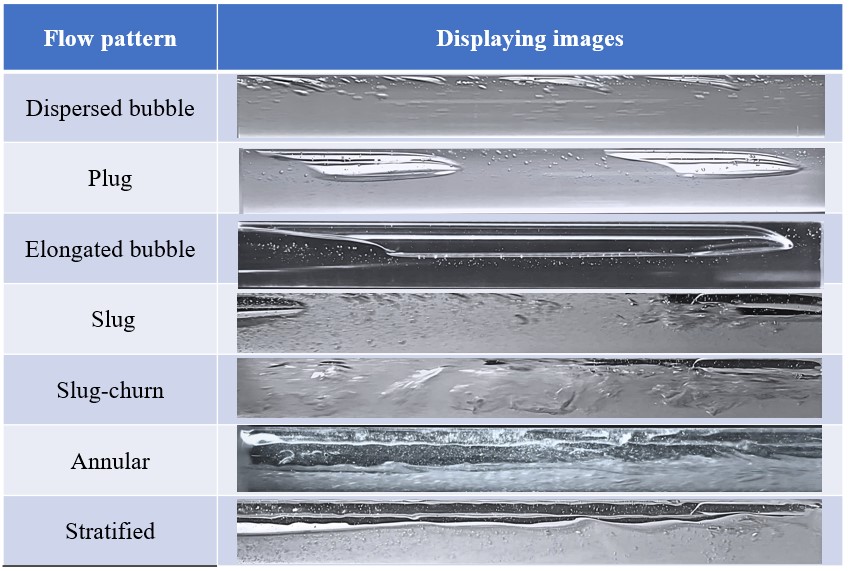}
\caption{High-speed camera images of flow patterns in the experimental flow rig.}
\label{fig:patterns}
\end{figure*}

The flow patterns observed in the transparent tube of the flow rig were captured using a combination of high-speed camera visualization and naked-eye observation, synchronized with a capacitance sensor. The transparent nature of the test section tube allowed for easy naked-eye observation of these patterns, particularly at lower flow rates, facilitating analysis of their development and transition from one pattern to another. The identified flow patterns formed on the flow rig are as follows:

\begin{enumerate}
\item The small bubble pattern in the experimental rig originated as a dispersed bubble flow near the inlet of the test section. These initially dispersed bubbles gradually merged and coalesced, transforming into a string of small bubbles located near the top of the pipe due to their buoyancy as shown in Figure~\ref{fig:patterns}. This phenomenon was observed consistently at all three angles of pipe inclination.

\item The plug flow pattern was observed exclusively at upward flow inclinations in the experimental rig. A slight increase in superficial gas velocity resulted in the formation of larger bubbles through the coalescence of smaller bubbles. This coalescence process led to the creation of cap-bubbles with a diameter approximately one-third that of the pipe and a length ranging from 1.5 to 2.5 times the pipe diameter as shown in Figure~\ref{fig:patterns}. However, it is important to note that the occurrence of the plug flow pattern was limited in extent within the experimental rig.

\item The elongated bubble flow pattern emerged with an increase in superficial gas velocity, resulting from the accelerated coalescence process between two or three adjacent plugs facilitated by the higher superficial gas velocity. This coalescence led to the formation of larger bubbles with a diameter slightly greater than half of the pipe diameter. The lengths of these bubbles ranged from 3.5 to 5.5 times the pipe diameter as shown in Figure~\ref{fig:patterns}. Notably, this flow pattern was observed at all three inclinations, with the longest bubbles observed in horizontal flow. As the angle of inclination increased, the length of the bubbles decreased.

\item Slug flow pattern is characterized by the liquid phase occupying approximately one quarter of the pipe diameter for the majority of the time. Subsequently, high amplitude waves begin to form, eventually accumulating to create a region where the liquid fills the cross-section of the pipe and is forcefully propelled by the air phase. These slugs, bridging the pipe, typically contain small discrete bubbles as shown in Figure~\ref{fig:patterns}. The length of each liquid slug falls within the range of 10 to 16 pipe diameters. This flow pattern is frequently observed in horizontal flow, while it becomes the dominant pattern at upward flow inclinations.

\item The slug-churn flow pattern occurs exclusively at high gas flow rates and exhibits similarities to slug flow. However, in this pattern, the liquid slugs undergo destruction, resulting in a transformation into a more foam-like and frothy state due to the increasing superficial gas velocity as shown in Figure~\ref{fig:patterns}. An oscillatory motion characterizes this flow pattern, making it challenging to estimate the length of the liquid slug-churn. Interestingly, the slug-churns travel through the pipe at a higher velocity than the liquid slugs, but with a lower frequency. This flow pattern is commonly observed in horizontal orientations and even more frequently in upward flow inclinations.

\item In annular flow, the gas phase exhibits a central flow along the core of the pipe, while the liquid phase forms a wavy film that adheres to the pipe wall. The two phases remain distinctly separated, and the thickness of the liquid film undergoes continuous fluctuations as shown in Figure~\ref{fig:patterns}. In horizontal flow, the liquid film tends to be thicker at the bottom compared to the top of the pipe. However, as the angle of inclination increases, the film thickness becomes more uniformly distributed around the pipe wall. This annular flow pattern is less frequently observed in horizontal orientations but becomes more prevalent with increasing angles of inclination.

\item In the stratified flow pattern, the gas and liquid phases remain fully separated within the pipe. The liquid phase flows along the bottom of the pipe, occupying approximately half of its diameter. Waves pass through the liquid phase but do not reach the top of the pipe wall due to insufficient amplitude as shown in Figure~\ref{fig:patterns}. This flow pattern is highly sensitive to the angle of inclination, as upward inclinations can lead to a transition to slug flow. At low flow rates, a transition occurs from stratified to slug flow, where only a few liquid waves touch the top of the pipe wall while the majority do not, resulting in a flow pattern that does not fit strictly into either the stratified or slug category.
\end{enumerate}

\subsection{Experiments on AI models}
\label{sec:exp-ai}
Based on the flow patterns observed in the experimental rig, AI-based classification models are being developed to automate the identification and classification of flow regimes. The models will use data collected from the capacitance sensor and the high-speed camera to train machine learning algorithms for pattern recognition. Preliminary results show that the combination of these data sources improves the accuracy of the model in identifying and predicting flow patterns under varying flow conditions.

 \begin{table*}[h]
\centering
\small
\begin{tabular}{>{\centering\arraybackslash}m{4cm}>{\centering\arraybackslash}m{2cm}>{\centering\arraybackslash}m{2cm}>{\centering\arraybackslash}m{2cm}>{\centering\arraybackslash}m{2cm}}
\toprule
\multirow{2}{*}{\textbf{Model}} & \multicolumn{2}{c}{\textbf{Experiment-Based }} & \multicolumn{2}{c}{\textbf{Pattern-Based}}  \\ 
& \textbf{Accuracy} & \textbf{F1 Score} & \textbf{Accuracy} & \textbf{F1 Score} \\ \midrule
Random Forest & 0.788 & 0.778 & 0.473 & 0.470 \\ 
Random Forest with PCA & 0.769 & 0.769 & 0.461 & 0.472 \\ 
SVM & 0.645 & 0.612 & 0.565 & 0.535 \\ 
SVM with PCA & 0.611 & 0.570 & 0.539 & 0.507 \\ 
MLP & 0.611 & 0.578 & 0.528 & 0.482 \\
MLP with PCA & 0.629 & 0.602 & 0.502 & 0.451 \\ 
Decision Tree & 0.638 & 0.639 & 0.412 & 0.428 \\ 
Decision Tree with PCA & 0.662 & 0.670 & 0.362 & 0.370 \\ 
1D CNN & 0.747 & 0.720 & 0.690 & 0.660 \\
Transformer & 0.706 & 0.680 & 0.676  & 0.660 \\ 
%Transformer with oversampling & 0.719 & 0.700 & 0.638 & 0.590 \\ \hline
1D SENet & \textbf{0.850} & \textbf{0.847} & \textbf{0.712} & \textbf{0.674} \\ \bottomrule
\end{tabular}
\caption{Results of AI models.}
\label{tab:eval_all}
\end{table*}

The best-performing results are highlighted in bold, and our 1D SENet achieves the best performance on both the experiment-based and pattern-based datasets. Although Random Forest performs relatively well on the experiment-based dataset, it experiences a significant drop in performance on the pattern-based dataset, with accuracy falling from over 78\% to only 47\%. This indicates that Random Forest is prone to overfitting, leading to poor generalization and out-of-distribution prediction performance.

Traditional machine learning models, when combined with PCA to reduce the data dimensionality to 15, show competitive performance compared to their versions without PCA, while also significantly reducing computational costs. However, these models generally exhibit poor generalizability, as evidenced by their substantial performance drops for the pattern-based dataset. In contrast, 1D SENet, 1D CNN, and Transformer models demonstrate superior generalizability, outperforming all traditional models on the pattern-based dataset. The 1D SENet achieves an accuracy of over 85\% on the experiment-based dataset, significantly outperforming other models. Although all models exhibit significantly lower performance on the pattern-based dataset, 1D SENet still achieves an accuracy that exceeds 71\%. This performance decrease is mainly due to the substantial change in the distribution between the experiments, such as the variations in means and standard deviations as described in Table~\ref{tab:file-list}, and the lack of more diverse experiments in the data set. These factors make it difficult to establish precise decision boundaries. Increasing the diversity and quantity of experiments, as well as employing data augmentation techniques, could significantly enhance performance and bring the results on the pattern-based dataset much closer to those of the experiment-based dataset.

\subsection{Hyperparameter Tuning}

\begin{table*}[t]
\centering
\footnotesize
\begin{tabular}{>{\centering\arraybackslash}m{3.5cm} >{\centering\arraybackslash}m{2.2cm} >{\centering\arraybackslash}m{2.2cm} >{\centering\arraybackslash}p{2.2cm} >{\centering\arraybackslash}p{2.2cm}}
\toprule
\multicolumn{5} {c} {\textbf{Hyperparameters \& Performance}}\\ \midrule
\textbf{Kernel Size} & 16 & 8 & 5 & 3  \\ [+0.1cm]
Accuracy & 0.821 & 0.796 & \textbf{0.821} & 0.749  \\
F1 Score & 0.809 & 0.786 & \textbf{0.816} & 0.754   \\ \midrule
\textbf{BN (Batch Norm) \& Dropout} & BN=False, Dropout=True & BN=True, Dropout=False & BN=False,  Dropout=False & BN=True, Dropout=True \\ [+0.1cm]
Accuracy & 0.756 & 0.730 & 0.804 & \textbf{0.821} \\
F1 Score & 0.701 & 0.717 & 0.797 & \textbf{0.816} \\ \midrule
\textbf{Dropout rate} & 0.5 & 0.4 & 0.3 & 0.2 \\ 
Accuracy & 0.821 & 0.831 & \textbf{0.849} & 0.834\\
F1 Score & 0.816 & 0.827 & \textbf{0.845} & 0.835\\ \midrule
\textbf{Width ratio} & 1 & 2 & 4 & 8  \\ [+0.1cm]
Accuracy & 0.722 & 0.813 & \textbf{0.821} & 0.717\\
F1 Score & 0.700 & 0.807 &  \textbf{0.816} & 0.680 \\ \midrule
\textbf{Block Stages} & 2 & 3 & 4 & 5 \\ [+0.1cm]
Accuracy & 0.750 & 0.821 & \textbf{0.850} & 0.834 \\
F1 Score & 0.764 & 0.816 & \textbf{0.847} & 0.831 \\ 
\bottomrule
\end{tabular}
\caption{Hyperparameter tuning of SENet.}
\label{tab:hyper}
\end{table*}
We conducted hyperparameter tuning for our 1D SENet to assess its sensitivity to key parameters and determine the optimal configuration. The most influential factors affecting the model's performance were identified, including the kernel size of the convolutional layers, the application of batch normalization and dropout, the dropout ratio (if applied), the width of the SENet, and the length, i.e., the number of block stages. The optimal kernel size was found to be 5, achieving the best results in terms of accuracy and F1 score. The combined use of batch normalization and dropout led to a noticeable improvement in performance. The model performed best with a width and block stage length of 4. Although the highest accuracy and F1 score were observed with a dropout ratio of 0.3, the overall best performance, when all factors were considered, was achieved with a dropout ratio of 0.5. As shown in Table~\ref{tab:hyper}, the performance of the model is highly sensitive to these hyperparameters, particularly the width, block stage length, and the use of batch normalization and dropout. Therefore, our final configuration for the 1D SENet and ResNet models includes a kernel size of 5, the use of both batch normalization and dropout with a dropout ratio of 0.5, and a width and block stage length of 4.

\begin{figure}[t]
\centering
\includegraphics[width=0.49\textwidth]{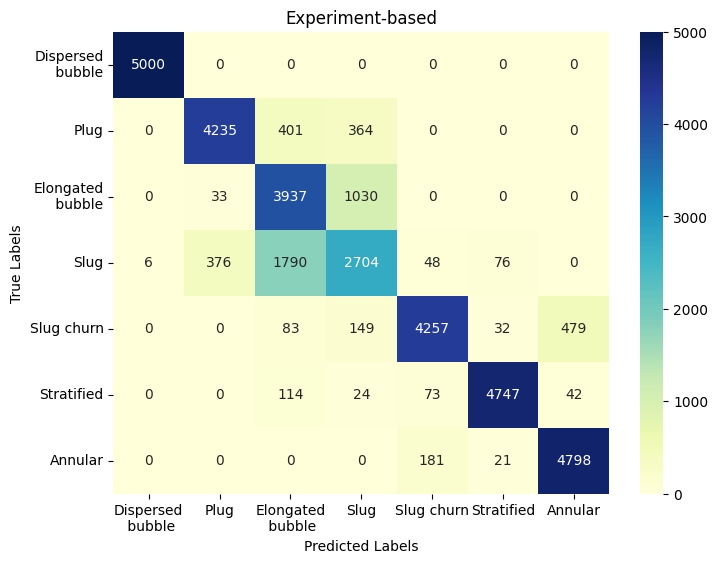}
%\vspace{0.5cm}  % Adjust space between images
\includegraphics[width=0.49\textwidth]{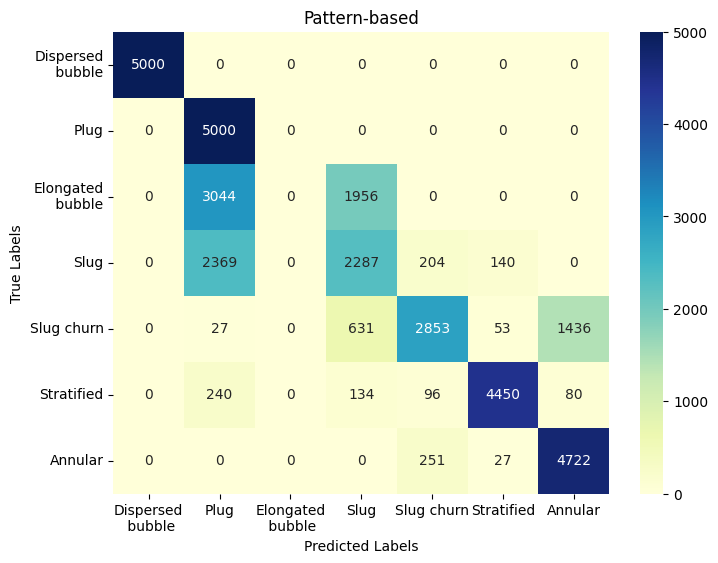}
\caption{Confusion matrix of SENet predictions: experimental-based (left) and pattern-based (right).}
\label{fig:confusion}
\end{figure}
% class_dict = {0:'small',1:'plug',2:'elon',3:'slug',4:'slug_churn',5:'stra_wavy',6:'annu_flow'} 
\begin{table}[h]
\centering
\footnotesize
\begin{tabular}{>{\centering\arraybackslash}m{2.cm} >{\centering\arraybackslash}m{3cm} >{\centering\arraybackslash}m{1.5cm} >{\centering\arraybackslash}p{1.5cm} >{\centering\arraybackslash}p{1.5cm}}
\toprule
\textbf{Category} & \textbf{Dataset} & \textbf{Precision} & \textbf{Recall} & \textbf{F1 Score} \\ \midrule
\multirow{2}{1cm}{Dispersed bubble} & Pattern-based & 1.00 & 1.00 & 1.00 \\
 & Experiment-based & 1.00 & 1.00 & 1.00 \\ [+0.1cm]
\multirow{2}{1cm}{Plug} & Pattern-based & 0.84 & 1.00 & 0.91 \\
 & Experiment-based & 0.91 & 0.85 & 0.88 \\ [+0.1cm]
\multirow{2}{1cm}{Elongated bubble} & Pattern-based & 0.00 & 0.00 & 0.00 \\
 & Experiment-based & 0.62 & 0.79 & 0.70 \\ [+0.1cm]
\multirow{2}{1cm}{Slug} & Pattern-based & 0.38 & 0.82 & 0.52 \\
 & Experiment-based & 0.63 & 0.54 & 0.58 \\ [+0.1cm]
\multirow{2}{1cm}{Slug churn} & Pattern-based & 0.62 & 0.49 & 0.54 \\
 & Experiment-based & 0.93 & 0.85 & 0.89 \\ [+0.1cm]
\multirow{2}{1cm}{Stratified} & Pattern-based & 1.00 & 0.79 & 0.88 \\
 & Experiment-based & 0.97 & 0.95 & 0.96 \\[+0.1cm]
\multirow{2}{1cm}{Annular} & Pattern-based & 0.77 & 0.83 & 0.80 \\
 & Experiment-based & 0.90 & 0.96 & 0.93 \\
\bottomrule
\end{tabular}
\caption{Precision, recall and f1 score of each category.}
\label{tab:prf1}
\end{table}

\subsection{Category Analysis}
We present the detailed precision, recall, and F1 scores of our SENet on two datasets, as shown in Table~\ref{tab:prf1}. Among the flow patterns, the dispersed bubble is the easiest to distinguish, achieving 100\% accuracy on both datasets. Plug, stratified wavy, and annular flows are relatively easier to classify, although the annular flow poses more challenges in the pattern-based dataset. The most difficult categories to classify are elongated bubble, slug, and slug churn, with the model completely failing to classify elongated bubble when the targets are out-of-distribution in the pattern-based dataset. However, when the test sets are within a distribution similar to the training set, the model achieves relatively acceptable accuracy for the elongated bubble and slug churn patterns. Slug remains one of the most challenging patterns to classify consistently, with F1 scores ranging between 50\% and 60\%.

\begin{figure}[h]
\centering
\includegraphics[width=0.49\textwidth]{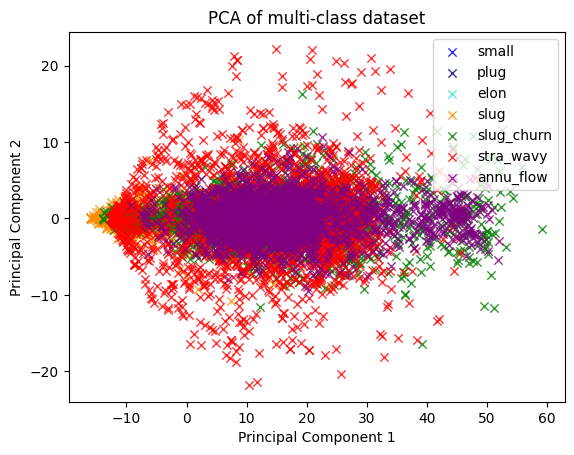}
%\vspace{0.5cm}  % Adjust space between images
\includegraphics[width=0.49\textwidth]{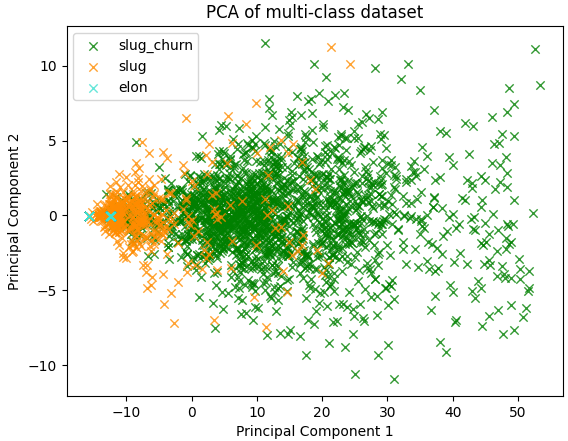}
\caption{PCA plotting of all classes and difficult cases.}
\label{fig:pca}
\end{figure}

To further assess the predictions of the model, we plot the confusion matrix for all classes, as shown in Figure~\ref{fig:confusion}. In the experiment-based dataset, where in-distribution performance is evaluated, the model frequently misclassifies slug as elongated bubble, with more than 35\% of slug samples being incorrectly classified. Additionally, more than 1,000 elongated bubble samples are misclassified as slug, indicating that these two categories are the easiest to confuse. This confusion is supported by the PCA visualization of the patterns in Figure~\ref{fig:pca}. In addition to this, plug and slug churn are the second most common sources of classification errors. In the pattern-based dataset, where the evaluation set's distribution differs from that of the training set, elongated bubble, plug, slug, and slug churn continue to be the most difficult to classify. Notably, none of the elongated bubble samples are classified correctly, as they are significantly overlapped with plug and slug. Other major errors include slug being misclassified as plug, and slug churn being misclassified as annular flow. These issues highlight areas for improvement in future work, particularly in addressing the high rate of slug misclassification as plug, with nearly half of the slug samples incorrectly classified.

\begin{figure}[h]
\centering
\includegraphics[width=0.49\textwidth]{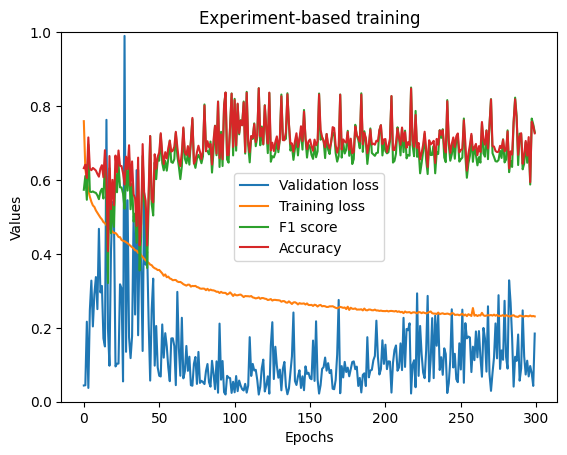}
%\vspace{0.5cm}  % Adjust space between images
\includegraphics[width=0.49\textwidth]{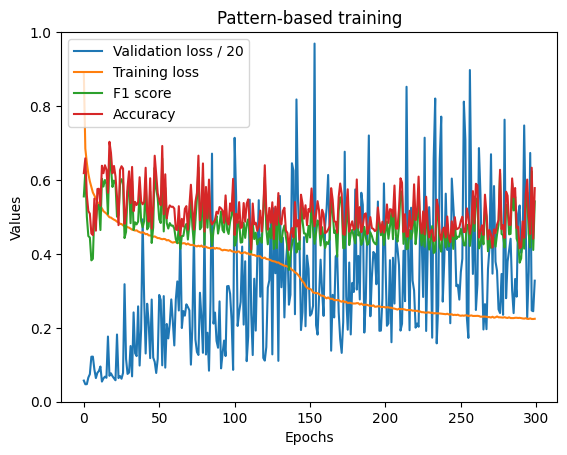}
\caption{Training diagrams of SENet on two datasets.}
\label{fig:training}
\end{figure}

\subsection{Training Analysis}
The training curves of SENet on the two types of datasets are visualized in Figure~\ref{fig:training}. The curves for the experiment-based dataset are significantly more stable than those for the pattern-based dataset, suggesting that more diverse flow pattern data can enhance the stability of training. However, after approximately 100 epochs, signs of overfitting emerge, with the model's accuracy starting to decline and the validation loss increasing.

In contrast, when training on the pattern-based dataset, the model tends to overfit more easily, and the validation loss becomes increasingly unstable in the later stages of training. This indicates that while the model has sufficient capacity to classify the data, it struggles with generalizability and stability when faced with data from different distributions. This instability highlights the need for further improvements to enhance the robustness of the model and its ability to generalize across diverse datasets.

\section{Conclusion and Future Work}\label{conclusion}
In conclusion, our models require only 5 seconds of data from two capacitance sensors, enabling real-time detection with a lightweight deep learning model (1D SENet) of just 12MB. By using a single modality of input -- signals from two sensors -- our approach offers significant advantages in terms of practicality, simplicity, real-time performance, and robustness in multi-phase flow pattern classification, outperforming previous methods that rely on multi-modal inputs, more than two sensors, and complex models and equipment.

We comprehensively applied AI models to classify flow patterns based on data obtained from two capacitance sensors. Our deep learning methods demonstrated strong performance and generalizability in distinguishing seven flow patterns, with our best model, 1D SENet, achieving over 85.0\% accuracy and an F1 score of 84.7\%. In a more challenging scenario where the evaluation set consists entirely of out-of-distribution data, 1D SENet still maintained an accuracy of 71.2\%, while traditional machine learning methods achieved a maximum accuracy of only 56.5\%.

We further analyzed the results through hyperparameter tuning, detailed precision and recall metrics, PCA visualization, confusion matrices, and training loss-accuracy diagrams. The analysis revealed that plug, elongated bubble, slug, and slug churn are the most challenging patterns to classify across both datasets, while dispersed bubble is the easiest to classify. These findings suggest future directions for improving the performance of AI models for the task of flow pattern classification, such as data augmentation, conducting additional experiments to create more diverse datasets, and refining model tuning to enhance generalizability.

\bibliographystyle{abbrvnat}
%\bibliography{References}

%Future work: more angles? more experiments, better performance
\newpage
\appendix
\section{Further Details on Experimental Design and Procedures}
\label{app1}
This appendix provides additional explanations and detailed descriptions of the experimental apparatus and methods utilized in our study. It elaborates on the design and calibration of the experimental setup, including the functionality and configuration of the capacitance sensor and the high-resolution video recording system. By offering in-depth insights into the materials, techniques, and procedures employed, this section aims to enhance the understanding of the research findings presented in the main paper. The information included here will aid readers in grasping the complexities of the experimental design and its significance.

A meticulously designed and calibrated experimental apparatus was employed to ensure the successful implementation of our experiment. This apparatus had been specifically customized to fulfill the precise demands of our research objectives, enabling variables of interest, such as the capacitance sensor and high-resolution camera recordings, to be accurately measured and controlled. In the following paragraph, a comprehensive explanation will be provided regarding the crucial components and functionalities of this apparatus, emphasizing its significance in relation to our experiment.

A test section measuring 6.0 meters in length was fabricated using an extruded transparent acrylic tube characterized by an internal diameter of 36.0 mm and an outer diameter of 40.0 mm. This construction was specifically designed to allow for visual assessment of flow dynamics. The inclination of the test section can be modified within a range of 0$^\circ$ to 35$^\circ$ due to the inclusion of a swinging table mechanism. Furthermore, the test section incorporates flexible tubes at its edges, facilitating the conduction of investigations and experiments at user-defined angles. To mitigate the influence of vibrations in the system, two jacks were utilized to support the test section. To supply the test section with fluid, oil was extracted from a tank with a capacity of 2 m$^3$. The oil was then pumped through a PVC pipe using a gear pump (Model GL-50-10). To prevent backflow, a check valve was installed in the supply pipe.

Gas injection was facilitated by the placement of an air compressor downstream of the liquid flow meter. Efficient homogenization of oil and gas was achieved through the implementation of a mixer device measuring 28.0 cm in length. This device consisted of an external pipe made of Perspex and an internal pipe made of brass, which featured numerous small holes. The inner pipe, which had the same diameter as the test section, was separated from the outer pipe by a 1.8 cm gap. It contained around 100 staggered holes, each measuring 1 mm in diameter, spaced 10 mm apart axially and 5.0 mm apart circumferentially, allowing air to enter smoothly. The oil entered in the axial direction, while air was introduced into the space between the two pipes and then entered the test section through the holes in the inner pipe, as shown in Figure~\ref{fig:setup} (c). After passing through the test section, the oil and gas were returned to the oil tank for recirculation. The gas naturally separated and evaporated into the surrounding area, while the oil was retained in the tank. This cyclic process continued consistently.

The experimental apparatus involved illuminating the test section to facilitate visual observation and recording of flow patterns on videotape. A transparent rectangular box filled with water, as shown in Figure~\ref{fig:viewbox}, was incorporated into the test section to mitigate image distortion caused by pipe curvature and to counteract the heating effect of the illumination by providing cooling. Flow characteristics were captured using an Analogue High-Speed Video system (CR4000 $\times$ 2, Optronis GmbH, Kehl, DE). This advanced camera enabled the acquisition of 500 frames per second for a duration of 10 to 20 seconds, resulting in a full resolution of 4 megapixels. Furthermore, the camera had the capability to achieve even higher acquisition rates for specific regions of interest within the flow field. Equipped with an F mount and a Nikon 50 mm f/1.8D lens, the camera ensured consistent and uniform lighting during the high-speed recordings by employing an LED assembly positioned behind the viewing box. Visual observations and photo-recordings were conducted at a distance of 5 meters from the inlet to ensure that the flow pattern had fully developed.

The capacitance sensor was installed at a distance of 5 m from the inlet within the test section, as shown in Figure~\ref{fig:sensor_geom}. It was securely affixed by clamping it around the test section in a curved configuration. The positioning of the excitation electrode at the top of the pipe facilitated precise detection of the flow pattern~\cite{al2020bsimplified, al2020systematic}. The sensor was divided into two parts to accommodate clamping, ensuring a tight fit around the external circumference of the pipe. The two electrodes were designed to always remain opposite each other. Moreover, an acrylic insulation material was used to fill the gap between the electrodes and the screen, providing mechanical stability, as shown in Figure~\ref{fig:sensor}. To minimize electrical interference, the insulation material was completely enclosed by a brass screen. This design offers several notable advantages, including portability and the absence of direct contact between the two electrodes and the fluid within the pipe. Consequently, issues such as erosion and disruption of flow patterns are effectively mitigated. To facilitate accurate measurements, the capacitance sensor was connected to a dedicated electronic capacitance measurement device. The output of the measurement device was then transmitted to an analogue-to-digital converter (ADC) via a shielded cable of minimal length to minimize noise interference. Both the capacitance sensor and the electronic devices were calibrated to operate within a capacitance range of 0 to 100 pF.

The capacitance measurement was carried out using an electronic device called the Capacitance Measurement Device (CMD), as shown in Figure~\ref{fig:sensor_geom}, which is equipped with a burst mode charge-transfer converter. The CMD has a Pulse Width Modulation (PWM) output with a specialized filter for gas/liquid flow applications. The device has eight-bit resolution for precise measurements and includes two calibration inputs for span calibration. The output of the CMD depends on both the load and the sampling capacitance. The CMD has an automatic compensation mechanism to minimize the impact of cable capacitance on measurement accuracy. It also has a control panel for easy calibration and LED indicators for displaying device status and errors. Capacitive sensors are designed to detect changes in electrical capacitance caused by variations in physical properties. These sensors typically have a narrow capacitance range, measuring a few picofarads or less. However, accurately measuring such low capacitance values is challenging because of external interferences such as noise and parasitic capacitances. Parasitic capacitances include cable capacitance and stray capacitance on PCB boards. These parasitic capacitances often exceed the intended measurements. To address these issues, techniques such as capacitance-to-voltage converters and capacitance-to-frequency converters have been proposed~\cite{smith1997scientist, ferry1997design, toth1996very}. The CMD compensates for cable capacitance to ensure measurement accuracy. It has a dedicated control panel for calibration and LED indicators for real-time information on device status and errors.

The Capacitance Measurement Device Capacitive Sensor Transducer is a specialized capacitance meter for capacitive sensors. It can measure capacitance values up to 24 pF and allows calibration within any subrange from 0 to 24 pF. This device can accommodate a wide range of capacitance measurement requirements for various applications.

In this study, a circuitry device and an analog low-pass filter were used in conjunction with a Pulse Width Modulation (PWM) output. This setup converted small capacitance values into a direct current voltage signal ranging from 0 to 5 V. The measurement principle, methodology, and specific details of this setup were based on pioneering work proposed in~\cite{ferry1997design, kollataj2008multi}. By adopting this configuration, numerous advantages were achieved, including:
\begin{enumerate}
    \item Minimal power consumption, ensuring efficient utilization of the battery power source.
    \item Reduced emission of radio-frequency signals, thereby enhancing electromagnetic compatibility.
    \item Elimination of the influence of electrical noise on the measurement system~\cite{smith1997scientist}.
    \item Rapid response with a frequency range extending up to 500 Hz.
    \item Improved circuit sensitivity and stability~\cite{ferry1997design, kollataj2008multi}.
\end{enumerate}

The calibration procedure used in this study involved a simple two-step process. The test section was first filled with air, and the corresponding capacitance value was measured and stored as 0 V. The test section was then filled with oil, and the resulting capacitance value was saved as 5 V. These calibration settings were retained as long as the device remained powered on, but were erased when the device was powered off, requiring recalibration each time it was turned on again.

\end{document}